\documentclass[lettersize,journal]{IEEEtran}
\usepackage{amsmath,amssymb, amsfonts}
\usepackage{algorithmic}
\usepackage{algorithm}
\usepackage{subfigure}
\usepackage{pdfpages}
\usepackage{tabularx}
\usepackage{array}
\usepackage{textcomp}
\usepackage{stfloats}
\usepackage{url}
\usepackage{verbatim}
\usepackage{color}
\usepackage{graphicx}
\usepackage{cite}
\usepackage{multirow}
\usepackage{colortbl}
\usepackage{hhline}
\begin{document}

\title{Federated Learning over a Wireless Network: Distributed User Selection through Random Access}

\author{Chen Sun,~\IEEEmembership{Senior Member,~IEEE,} 
Shiyao Ma, 
Ce Zheng,
Songtao Wu,
Tao Cui,
Lingjuan Lyu
}



\maketitle

\begin{abstract}
User selection has become crucial for decreasing the communication costs of federated learning (FL) over wireless networks. However, centralized user selection causes additional system complexity. This study proposes a network intrinsic approach of distributed user selection that leverages the radio resource competition mechanism in random access. Taking the carrier sensing multiple access (CSMA) mechanism as an example of random access, we manipulate the contention window (CW) size to prioritize certain users for obtaining radio resources in each round of training. Training data bias is used as a target scenario for FL with user selection. Prioritization is based on the distance between the newly trained local model and the global model of the previous round. To avoid "excessive contribution" by certain users, a counting mechanism is used to ensure fairness. Simulations with various datasets demonstrate that this method can rapidly achieve convergence similar to that of the centralized user selection approach.

\end{abstract}

o
\section{Introduction}
Artificial intelligence (AI) has gained widespread popularity as a technology that is used to enhance various aspects of our daily lives and augment human capabilities. For instance, in vehicles, AI engines can identify objects such as cars or pedestrians with front-facing cameras, even in low-light conditions where the human eye may fail. However, training an AI model for such tasks requires the collection of extensive image data from diverse locations. Collaborative training of an AI model on a cloud server by multiple users can facilitate training with a wider range of scenarios, but this requires uploading significant amounts of image data, which raises privacy concerns~\cite{lyu2022privacy}.

To address the concerns related to user privacy and mitigate the heavy traffic load associated with uploading training images, federated learning (FL) through wireless networks has emerged and is a widely used approach for autonomous driving scenarios \cite{Lu2022BC_FL}. In this system, each user trains a model locally and transmits only the model parameters, rather than the images, to the cloud via a wireless network. Through a downlink channel, an updated model that incorporates the local updates from all the users is generated. This process iterates until the model's performance converges.

However, the application of FL over wireless networks faces the challenge of limited wireless resources for model uploading. One solution overcomes this issue is to choose a subset of users in each round of the FL process. Consequently, user selection has become a crucial area of research in FL. User selection is normally based on certain rules to improve the performance of FL in the following ways. 

First, training data are critical to the FL process. When cars that are located close to each other use their front cameras to gather training data, the resulting data may have correlations due to the overlapping camera views. Using these data can lead to a decrease in the training performance. To address this issue, it was recommended in \cite{sun2022user} that users who are sufficiently distant from each other be selected in order to reduce correlation in the data. Moreover, the assumption that training data collected by all users are independent and identically distributed (IID) may not always be accurate. This is because the training data come from a diverse group of users, each with unique preferences, such as preferences for driving on highways or city roads. Consequently, these preferences may result in biases within each user's data. In their work \cite{pmlr-v151-jee-cho22a}, Cho et al. put forth a user selection strategy where the server chooses users with the highest loss out of all users. 

The second point pertains to the uniqueness of the device of each user involved in the training process, including computational power and the algorithm employed in local training. Users with varying computational capabilities conduct their local training at different times during the FL process, and waiting for the slowest user to finish their training is not an efficient approach. To mitigate this issue, researchers proposed a solution in \cite{Zhengce23_FL_straggler}; in this approach, users with similar completion times are selected in each round of training.

Third, the success of FL relies heavily on the accessibility and reliability of radio resources, particularly when transferring models from users to the server. As a result, FL researchers are interested in user selection schemes that consider user radio resources. In \cite{Howard_Tony_FL_wireless}, the authors investigate the convergence of FL with existing wireless network scheduling policies for users. In \cite{DT_FL_SunWen20}, users with high communication capacity and computing capacity were encouraged to participate the FL process. With the digital twin modeling of these physical parameters, incentive schemes for the FL process were designed.


Motivated by the findings mentioned above, the telecommunication industry has undertaken the development of cellular network standards to enable the widespread adoption of FL over wireless networks. These standards facilitate the selection of users based on predetermined rules by providing an application programming interface (API) of its core network functions to the FL server, thereby creating an open platform for FL \cite{23.700-80}. As an illustration, the 5G network's core network can provide an FL application server with users' locations, enabling the server to handpick certain users. Additionally, the FL server can stipulate specific user selection criteria at network functions, and subsequently, the 5G core network functions, e.g., the network exposure function (NEF), can present a roster of users filtered by the selection criteria.

These core network functions were originally designed to allow cellular network operators to control and manage wireless connections of users including access control, roaming, quality of service (QoS) management, etc. Improving these functions to support FL application over wireless networks is limited by the inherent design philosophy of legacy wireless communication protocols, which prioritize the efficiency of information transportation rather than computational objectives. Selecting users for FL involves collecting their training characteristics, which can result in increased communication and computing overhead due to additional parameter uploads and radio resource control. Additionally, user selection and resource allocation of wireless resources typically rely on a centralized mechanism.


With the increasing demand for widespread connectivity, leveraging unlicensed bands and deploying wireless networks more flexibly and densely is becoming increasingly crucial. Communication relies on a random access mechanism by which each user competes for radio resources once he or she has data to transmit. This trend emphasizes the importance of decentralizing radio resource management. To enable the deployment of FL over these wireless networks, distributed user selection becomes necessary. In this regard, we propose the manipulation of the wireless resource competition mechanism based on the unique characteristics of each user's local learning process to achieve user selection with random access. As an example, we modify the carrier sensing multiple access (CSMA) mechanism by reducing the contention window (CW) size for selected users. This adjustment provides users with a greater chance to gain access rights and upload their local AI models to a server. Taking data bias as a target scenario for user selection, an issue arises, namely, how to determine the CW size in order to select users while avoiding the training data bias issue. 


\section{Preliminaries}
\subsection{FL with User Selection}
\label{sec: federated learning}
FL is a machine learning framework that operates in a distributed manner, it involves two key entities: the server and users. This approach enables the collaborative training of a global model for completing common tasks without directly accessing a user's data. The training process involves three key steps:
\begin{itemize}
    \item \textbf{Broadcast the global model:} At the start of each training round, the server, which is normally deployed on the cloud or edge server, selects a subset of users $\mathcal{K}^t$ to participate in the training process. The server then broadcasts the aggregated global model to these users through a wireless network, with the exception of the first round during which the server broadcasts the initialized model to all users. 
    \item \textbf{Local training:} Upon receiving the global model, users proceed to train it on their respective local datasets, resulting in the creation of local models. This process can be expressed as $\omega_k^t = \omega^t - \eta \nabla F(\omega^t)$, where $\omega^t$ is the global model in the $t$-th round. Here, $\omega_k^t$ is the local model of the $k$-th user in the $t$-th round and $F$ is the objective function. Subsequently, the users upload their local models to the server over wireless connections.
    \item \textbf{Aggregating the global model:} The server aggregates the local models by using FedAvg \cite{mcmahan2017communication} to generate the new global model that is expressed as follows:
    \begin{equation}
        \omega^{t+1} =   \frac{\| D_{k} \|}{\sum_{k \in \mathcal{K}^t}\| D_{k} \|},
    \end{equation}
where $D_k$ is the local dataset of the $k$-th user.
\end{itemize}
This process is repeated until convergence of the global model is achieved.

\subsection{FL with Random Access}

The evolution of wireless connectivity requires flexible use of radio spectrum such as the use of unlicensed bands, for example the WiFi system where radio resources as media for connection are obtained through a competition mechanism. A device must first determine whether media are being used by someone before sending signals. If the media are in use, the device enters a backoff state and waits for a specific period before attempting to transmit again \cite{LET_WiFi_Backoff_2020}. This backoff time is a randomly selected number of time slots (a 20 $\mu$s duration), and the random number must be within the value of the CW. Consider the scenario where the FL is implemented over a WiFi network and the server is implemented on the access point (AP) or on the cloud that is connected to this AP. Local training is performed by users who are connected to the FL server through a WiFi network. The wireless channel resource by which each user uploads their local model in each round depends on a competition mechanism. In the implementation described in \cite{FedFly}, FL is executed on Raspberry Pi devices connected via a WiFi network. Each user has an equal opportunity to upload their local model but in a random fashion.

The Third Generation Partnership Project (3GPP) cellular system also defines a mode for sidelinks by which devices that communicate directly with another need to obtain radio resources from a region-based resource pool. In such a system a device performs signal strength measurements on radio resources over a certain time window. If the energy is higher than a predetermined threshold, a channel is considered busy. The transmission parameters are subsequently set based on the channel busy ratio (CBR) \cite{NR_V2X_R16R17}. A typical use scenario where FL is implemented for autonomous driving applications utilizes edge-based FL server communication with cars via cellular-based sidelink communication protocols.

When implementing FL over an operator's 5G network, radio resource allocation can be centrally controlled for model transfer. However, when FL applications are executed such that users and the server communicate with a random access mechanism, the subset of users participating in each round of the learning process, denoted as $\mathcal{K}^t$, becomes randomly dependent on this autonomous resource allocation for each user. To address user selection and avoid system complexity with centralized resource allocation, we propose studying user selection for FL with random access. We suggest prioritizing certain users by adjusting their resource competition mechanism based on learning properties to increase the probability of obtaining available channels. For example, in the WiFi system we can modify the CW size and in the sidelink we can modify the threshold in the signal strength measurement, thus changing the CBR of the resources. The priority level information of each user that is broadcasted to other users who are competing for resources in the sidelink control information (SCI) can also be adjusted.

\section{Prioritized Radio Resource Competition}

This paper uses WiFi as a case study to demonstrate how to adjust the resource competition according to priority level of each user. The data bias scenario is taken as a target application area where user selection is critical to FL performance. Compared to the centralized approach, the proposed method only requires the transmission of locally trained models to the server without extra parameters related to the training process by the users.

The aggregating step, which involves collecting knowledge from each user, is a crucial part of the FL training process, as highlighted in Sec.\ref{sec: federated learning}. The level of contribution from each user is determined by the distance between the newly trained local model and the aggregated model, with a greater distance indicating a more significant contribution. To leverage this insight, we propose a user selection approach that prioritizes users with greater distances from the global model for the data bias scenario. Note that other learning characteristics such as computation power and battery life can be utilized to adapt the resource competition mechanism in different application scenarios. Here, we adopt the distance calculation method proposed by \cite{bernstein2020distance} to compute the distance between the two models, i.e., the newly trained local model and the global model. We define the priority of a user using the following formula:
\begin{equation}
    \text{priority} = \prod\limits_{l=1}^L (1 + \frac{\parallel \omega_{k,l}^t-\omega^t_l \parallel_2 }{\parallel \omega^t_l \parallel_2})\text{,}
    \label{eq: priority}
\end{equation}
where $L$ is the number of layers in neural network, $\omega^t_l$ is the $l$-th layer in the global model of the $t$-th training round, and $\omega_{k, l}^t$ is the $l$-th layer in the local model of the $k$-th user. Note that we have checked the values of the priority level in a few experiments and found that it is normally within [1, 1.2] and is not affected by different models. We adjust the CW size, $W$, through the priority of users as follows. Then, we generate a random number, $R$, that is uniformly distributed within (0, 1) to multiply the window size $W$ to obtain the backoff time, $T_{backoff}$ as follows:
\begin{equation}
        W = \frac{N}{\text{priority}}, \text{   }T_{backoff} = R * W,
\label{eq: backoff time}
\end{equation}
where $N$ is the hyperparameter to control the range of common CW sizes as the basis for adjustment. After completing their local model training, users upload their models to the server through CSMA with their own backoff time $T_{backoff}$. Those users with lower values of $W$ are likely to upload their models earlier than those with higher values of $W$. When the FL server is configured to merge the uploads of the first few users and broadcast the updated model, user selection is achieved through this backoff mechanism. It is worth noting that this user selection method is influenced by the order in which users upload their local models.

It is possible that user selection will become biased, with the server more likely to aggregate with users that have higher priority in the upload process. This phenomenon is particularly apparent in the non-IID scenario, where some categories of data are difficult to train or exclusively owned by certain users. As a result, there is a significant variation between the newly trained local models of these users and the global model, causing the server to select these users with higher frequency or probability. This bias toward certain users can cause the global model to become biased as well. To mitigate this, we propose that each user maintains a counter to adjust the distribution of user participation in training, thus reducing the impact of this bias on the final global model. To adjust the distribution of user participation in training, before uploading their local models to the server, users first check their selected frequency using their counters. If the number of times a user has been selected exceeds a certain threshold, they do not upload their local models. This process limits the frequency of user selection via the use of their counters. The counter for each user is calculated as a percentage of the number of times that a particular user has contributed to the global model divided by the total number of users that have contributed to the global model. Note that if the server of the FL is configured to merge $\mathcal{K}^t$ in each round, after $T$ rounds the server has merged models contributed by users $\sum_{t=1}^T\mathcal{K}^t$ times. Suppose one user has obtained channels and uploaded their local model $k$ times. In this case, the counter value for this user would be $k/\sum_{t=1}^T\mathcal{K}^t$ in the $T$-th round.


\begin{figure}[ht]
	\centering
	\large
        {\includegraphics[width=0.9\linewidth]{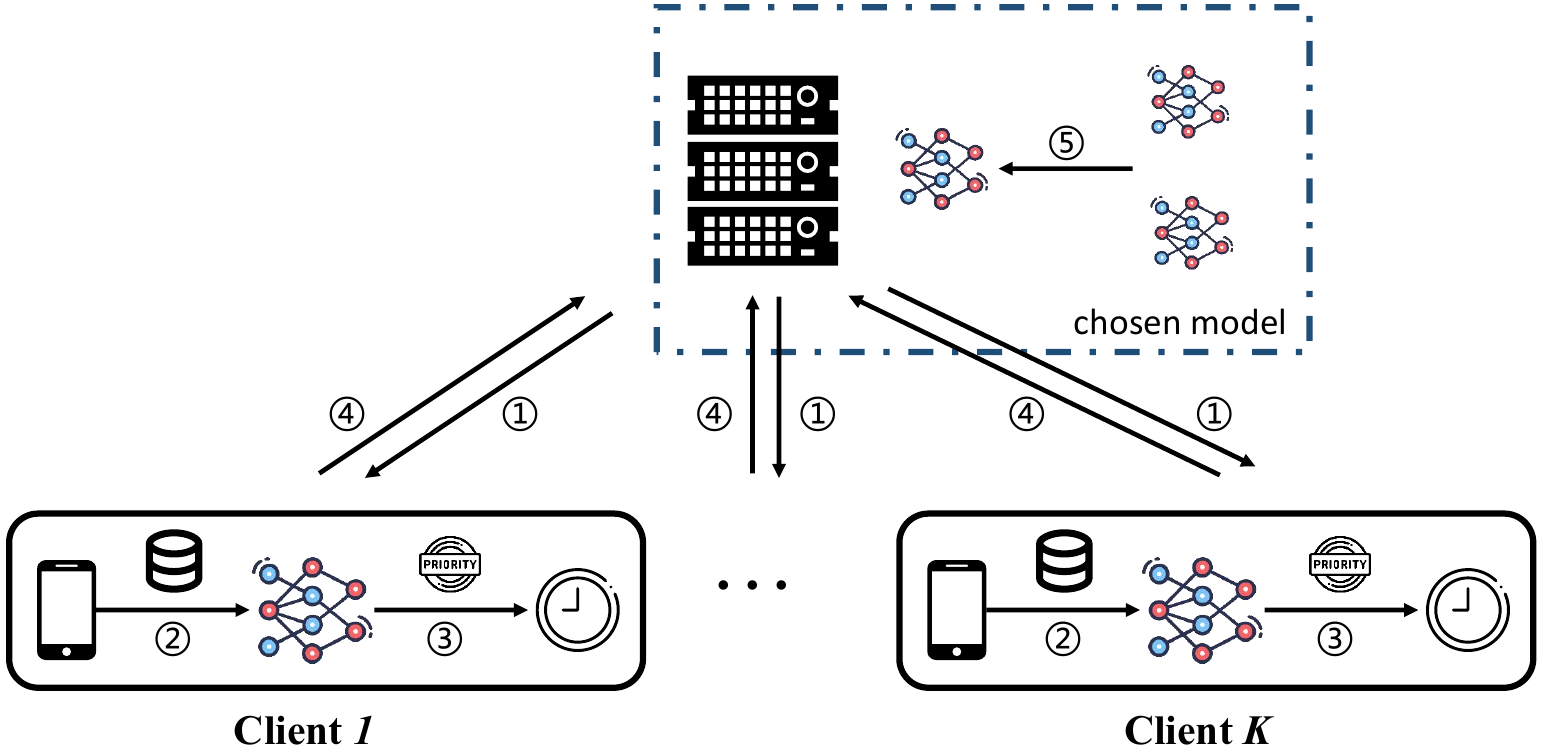}}
	\hfill
        \caption{Training procedure of prioritized channel access.}
        \label{fig: decentralized framework}
\end{figure}

In summary the process of FL with distributed user selection through random access is shown in Fig.\ref{fig: decentralized framework}. There are five steps: 
\begin{itemize}
    \item Step 1: The server broadcasts the global model that was aggregated in the previous round of training to the users. 
    \item Step 2: The users train the global model on their local datasets to generate local models.
    \item Step 3: The users calculate their priority with  Eq.~(\ref{eq: priority}), and then obtain the backoff time $T_{backoff}$ with Eq.~(\ref{eq: backoff time}).
    \item Step 4: Users check their counter values $k/\mathcal{K}^t$ and if their values exceed the threshold, they refrain from uploading their local models. Otherwise, they upload their local models according to the backoff time $T_{backoff}$ obtained in Step 3.
    \item Step 5: After receiving a certain number $\mathcal{K}^t$ of local models, the server applies the FedAvg algorithm to aggregate the local models and generates the updated global model. Subsequently, the server broadcasts this global model to all users, which also indicates that the server will no longer receive any more local models in this round. Then, each user updates their local counter value. Finally, after uploading, the users update their counter values as follows: if the upload was successful, they increase the numerator by one and the denominator by $\mathcal{K}^t$; if the upload was not successful, they only increase the denominator by $\mathcal{K}^t$ in their counter.
\end{itemize}
The framework repeats these steps until the model converges.

\section{Numerical results}
\subsection{Experiment setup}
In this section, we present the results of our experiments on two widely used datasets, Fashion-MNIST and CIFAR-10, to assess the performance of our proposed framework.

\subsubsection{Data partition} We conducted experiments under two scenarios: IID and non-IID. In the IID scenario, we randomly partitioned the datasets into equal parts, with each user receiving an equal share. In the non-IID scenario, we adopted the partitioning method proposed in \cite{mcmahan2017communication}, where the dataset is sorted by real label and partitioned into 200 shards of size 300, with each user receiving two shards. This results in each user having examples of only two types.

\subsubsection{Training settings} We set the number of users to $K=10$, and in each round, the server selects two users to participate in the training process. We use two neural networks, multilayer perception (MLP) and convolutional neural network (CNN), for image classification, and these networks are commonly used by researchers \cite{zhu2019multi}. The MLP consists of one hidden layer with 200 nodes, and its size is $d_{input} \times 200 \times 10$, where $d_{input}$ is the input dimension ($784$ for Fashion-MNIST and $3072$ for CIFAR). The CNN has two convolution layers with $5 \times 5$ kernels (128 channels and 256 channels, respectively) and a fully connected layer with 4096 nodes for Fashion-MNIST and 6400 nodes for CIFAR. Its size is $C_{input} \times 5 \times 5 \times 128 + 128 \times 5 \times 5 \times 256 + d_{fl} \times 10$, where $C_{input}$ is the input image channel ($1$ for Fashion-MNIST and $3$ for CIFAR) and $d_{fl}$ is the dimensions of the fully connected layer ($1024$ for Fashion-MNIST and $3072$ for CIFAR). We use stochastic gradient descent (SGD) to train the model parameters. The learning rate is set to $10^{-2}$, the batch size for training and testing is 32, and the local epoch is 1. We use FedAvg as the aggregation algorithm \cite{mcmahan2017communication}. Furthermore, we set $N$ to 2048 as the base to control the user window size based on priority levels and set the threshold to restrict the selected frequency of users to $16\%$ of the total selected frequency.

\subsubsection{Baseline} To assess the efficacy of our proposed technique for enhancing framework performance we selected the conventional user selection method, namely random selection, as our baseline. Additionally, we have considered both centralized and distributed user selection. The performance of the IID dataset is also shown as a reference.

\begin{figure}[h]
	\centering
	\large
	\subfigure [Fashion-MNIST]{\includegraphics[width=0.44\linewidth]{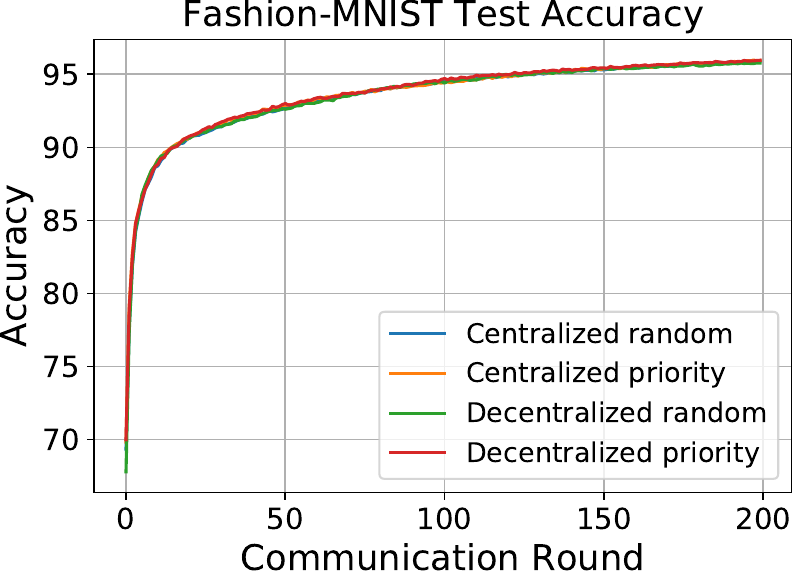} \label{fig: fmnist_iid}}
	\subfigure [CIFAR] {\includegraphics[width=0.45\linewidth]{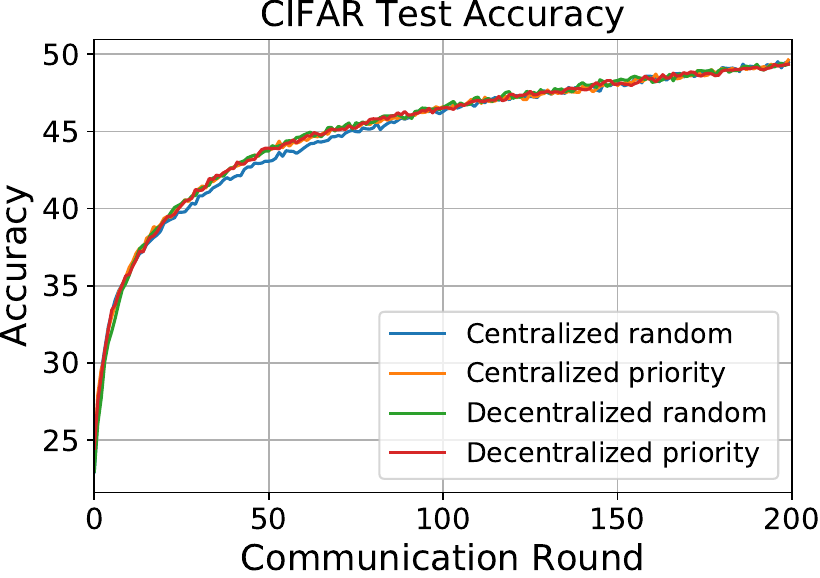} \label{fig: cifar_iid}}
	\hfill
        \caption{Comparison of four user selection strategies on two datasets.} 
        \label{fig: iid}
\end{figure}
\subsection{User selection with the IID dataset}
Initially, we examined our proposed technique in the IID scenario. In Fig.~\ref{fig: iid}, it is evident that the performance of the random selection and priority-based strategies is comparable in both datasets. This is explained by the similarity in data distribution among users, which results in comparable distances between each local model and the global model. Consequently, the likelihood of each user participating in the priority-based strategy is similar to that of each user participating in the random selection strategy. As a result, the performance of the four user selection strategies is comparable.

\subsection{User selection with non-IID dataset}
\begin{figure}[!h]
	\centering
	\large
	\subfigure [MLP]{\includegraphics[width=0.45\linewidth]{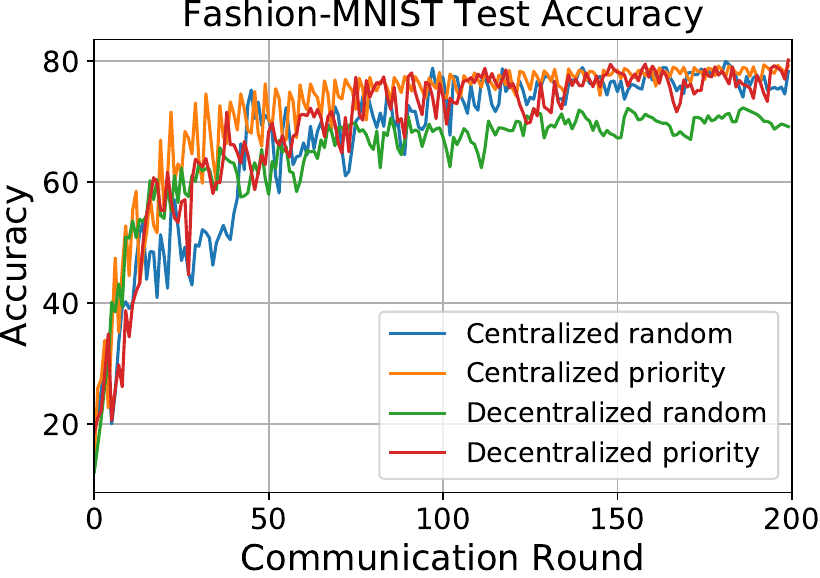} \label{fig: fmnist_non_iid}}
       \subfigure [CNN]{\includegraphics[width=0.45\linewidth]{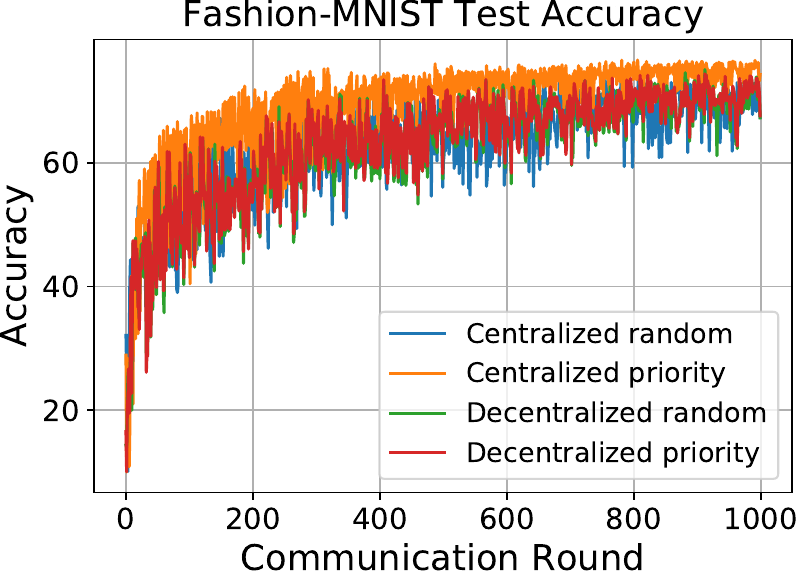} \label{fig: fmnist_non_iid_cnn}}
	\subfigure [MLP] {	\includegraphics[width=0.45\linewidth]{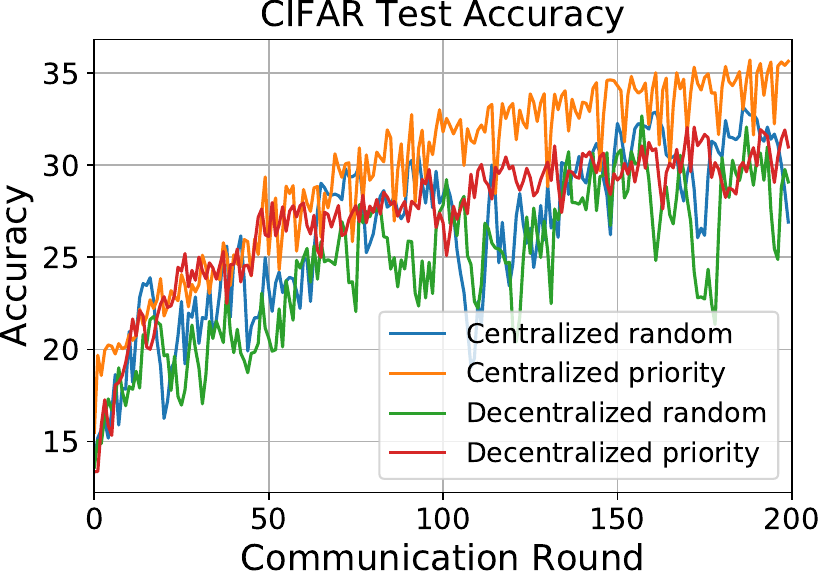} \label{fig: cifar_non_iid}}
       \subfigure [CNN] {	\includegraphics[width=0.45\linewidth]{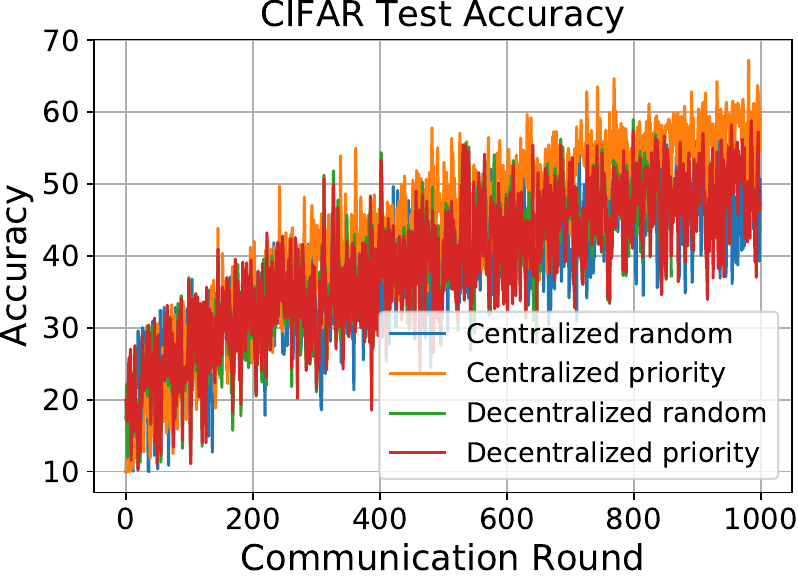} \label{fig: cifar_non_iid_cnn}}
	\hfill
        \caption{Comparison of four user selection strategies on two datasets, with the counter threshold set to $16\%$ and $N=2048$.}
    \label{fig: non-iid}
\end{figure}

To evaluate the effectiveness of the proposed method in the non-IID scenario, we conducted experiments on the two datasets with two networks, as shown in Fig.~\ref{fig: non-iid}. It is evident that the performance of centralized user selection and decentralized user selection is similar when users are selected randomly. However, by utilizing the priority strategy proposed in this paper, we observe an improvement in performance compared to that achieved by random user selection. This is because users are selected based on the distance between local models and the global model, with users that are farther away from the global model contributing more knowledge than others. Additionally, it is shown that decentralized user selection with random access inferiors is the centralized priority due to backoff time control, which alters the likelihood of a selected user acquiring the channel before others. However, there remains some uncertainty in winning the channel during the competition process. Notably, in Fig.~\ref{fig: non-iid}(a), selection with decentralized resource allocation can achieve performance similar to that of the centralized priority approach with low system overhead.

The simulation shows that while the trend aligns with our expectations, the performance of FL with the CNN varies greatly. This can be explained by the fact that our training data are highly biased and we select a small number of users in each round of training. Furthermore, the substantial volume of parameters of the CNN results in significant changes in model distance values and corresponding variation in the knowledge contributed by each user. As a result, we have decided to focus on the MLP architecture with the Fashion-MNIST dataset for our subsequent experiment.

\subsection{Fairness}
\begin{figure}[h]
    \centering
    \includegraphics[width=0.86\linewidth]{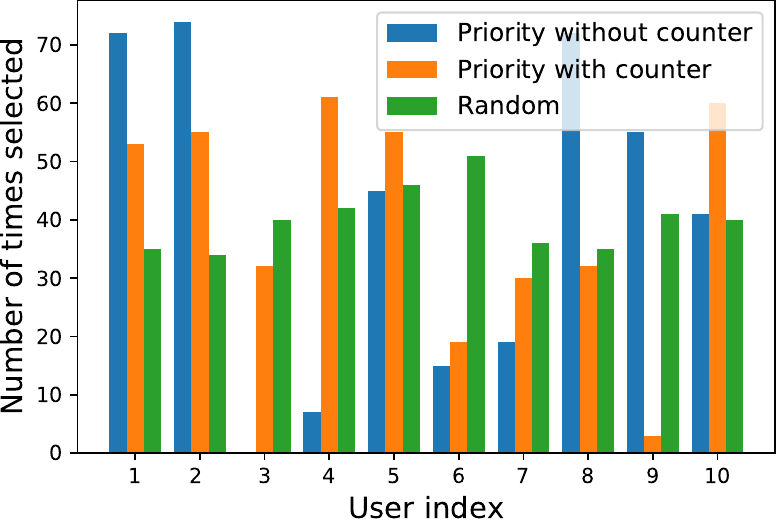}
    \caption {Comparison of the times being selected in the centralized scenario.}
    \label{fig: freq_counter}
\end{figure}

\begin{figure}[h]
        \centering
         \includegraphics[width=0.87\linewidth]{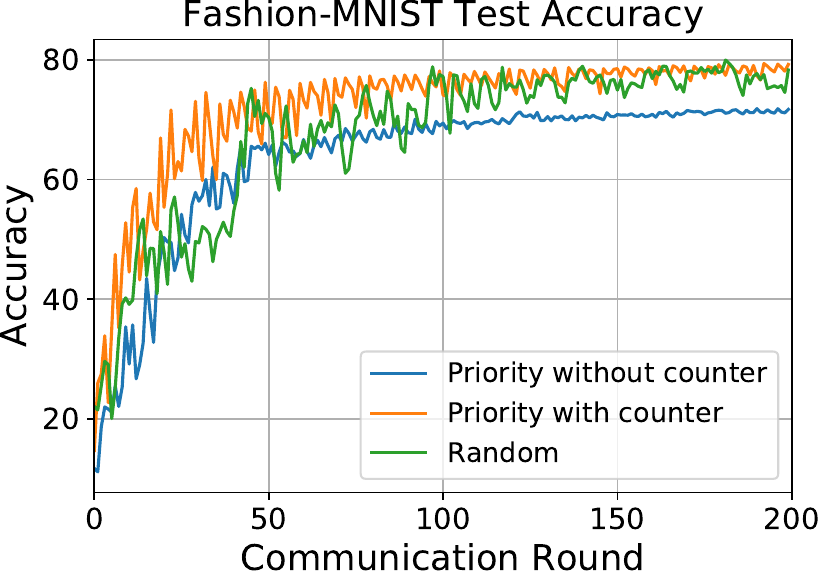}
       \caption{Performance with and without a counter in the centralized scenario.}
        \label{fig: freq}
\end{figure}

In this subsection, we focus on the fairness of the user selection process, with a specific emphasis on studying the effect of using a counter in the process of user selection for non-IID datasets. To isolate the effect of the uncertainty introduced by the random access, we utilize the centralized approach to assess the impact of the counter. We included ten users, each with two randomly selected shards of data, in the FL process, and two users were selected in each round. Fig.~\ref{fig: freq_counter} displays the number of times each user is selected. For random selection, each user is selected in relatively equal measure. In contrast, when users are selected based on priority without a counter, the selection is biased toward certain users, such as users 1, 2, and 3. We conducted several experiments and discovered that the most frequently selected users are generally associated with training data with digital labels, such as 2, 5, 8, and 9, corresponding to pullover, sandal, bag, and ankle boot image datasets, respectively. Additionally, the biasing phenomenon is more pronounced when the training datasets of a user correspond to 2 and 9 than other situations. We speculate that the features of these training images differ significantly from those of others and have been learned by the AI model, even though these features might be difficult for the human eye to detect. As a result, these models have different knowledge from other models and have a large model distance. The biased selection of certain users decreases the detection performance, as shown in Fig.~\ref{fig: freq}.

 If the counter is used, we can see that the number of times each user is selected becomes more balanced in Fig.~\ref{fig: freq_counter}, which ensures the effectiveness of our proposed method and mitigates the biasing global model over a few users. Consequently, the corresponding performance in Fig.~\ref{fig: freq} improves significantly and outperforms the random selection approach. The threshold value of the counter also influences the performance, and we carried out experiments with different values. Based on the results, we set the threshold to $16\%$.

\subsection{Effect of CW size}
Finally, we conducted experiments to investigate the impact of the hyperparameter $N$ in Eq.~(\ref{eq: backoff time}) of CW on the performance of the system. We varied the base of CW size $N$ from 512 to 2048 and observed that increasing the window size improves the performance. This can be attributed to the fact that a smaller window size often results in users producing similar backoff times, reducing the effectiveness of the prioritization mechanism for radio access. A larger window size promotes greater diversity in backoff times across users, making the prioritization mechanism more effective. Furthermore, the optimal CW size depends on the threshold values of the counter used in the selection process. These two parameters shall be adjusted for different scenarios.
\begin{figure}[h]
    \centering
    \large
    \includegraphics[width=0.85 \linewidth]{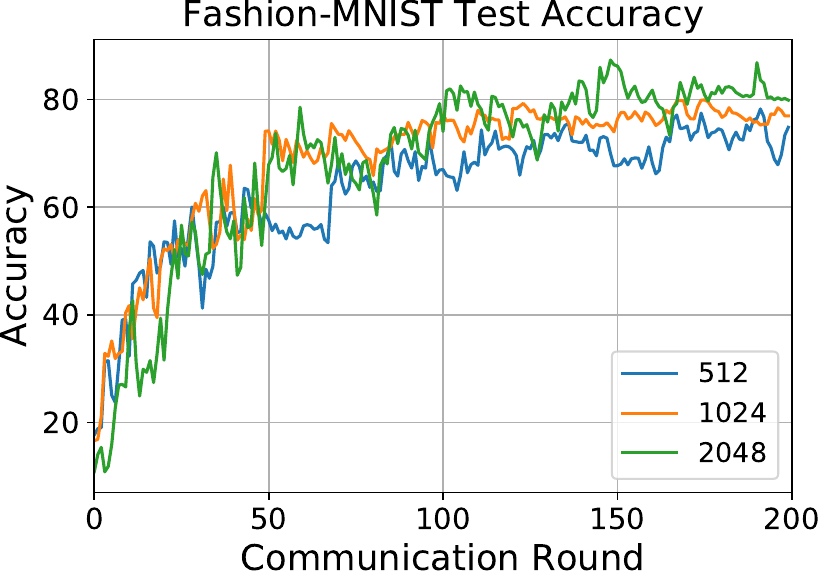}
    \hfill
    \caption{Comparison of different window sizes.}
    \label{fig: window_size}
\end{figure}

\section{Conclusion}

The aim of this study is to investigate the user selection process in FL, where users upload their locally trained models to the server with random access over a wireless channel. We introduced a user selection scheme by manipulating wireless resource competition to prioritize certain users. Different user selection criteria can be used to set user priority levels. In this paper, we considered the user selection problem in the context of training data bias and used CSMA as the random access scheme. Users are selected based on the knowledge they can contribute to the global model, which is measured by the distance between their local model and the global model. Prioritization was realized by reducing the CW size for selected users. To prevent biasing of the global model toward certain users, we also implemented a counter for each user to limit the number of their contributions. Our results demonstrated that the proposed prioritization strategy in the resource competition process, along with a counter, can improve the performance of FL. This allows FL with random access over wireless channels to achieve satisfactory performance while avoiding the complexity issue of the centralized approach.

\bibliographystyle{IEEEtran}
\bibliography{ref}

\end{document}